\documentclass[conference]{IEEEtran}
\IEEEoverridecommandlockouts
\usepackage{cite}
\usepackage{amsmath,amssymb,amsfonts}
\usepackage{algorithmic}
\usepackage{graphicx}
\usepackage{textcomp}
\usepackage{booktabs}
\usepackage{xcolor}
\usepackage{url}
\def\BibTeX{{\rm B\kern-.05em{\sc i\kern-.025em b}\kern-.08em
    T\kern-.1667em\lower.7ex\hbox{E}\kern-.125emX}}
\begin{document}

\title{\textsc{LoST}: A Mental Health Dataset of Low Self-esteem in Reddit Posts
}

\author{\IEEEauthorblockN{Muskan Garg}
\IEEEauthorblockA{
\textit{Mayo Clinic}\\
Rochester, MN, USA \\
garg.muskan@mayo.edu}
\and
\IEEEauthorblockN{Manas Gaur}
\IEEEauthorblockA{
\textit{University of Maryland, BC}\\
Baltimore, MD, USA \\
manas@umbc.edu}
\and
\IEEEauthorblockN{Raxit Goswami}
\IEEEauthorblockA{ 
\textit{HealthcareNLP LLC.}\\
Louisville, KY, USA \\
raxit.G@healthcarenlp.com}
\and
\IEEEauthorblockN{Sunghwan Sohn}
\IEEEauthorblockA{
\textit{Mayo Clinic}\\
Rochester, MN, USA \\
sohn.sunghwan@mayo.edu}
}

\maketitle

\begin{abstract}
Low self-esteem and interpersonal needs (i.e., thwarted belongingness (TB) and perceived burdensomeness (PB)) have a major impact on depression and suicide attempts. Individuals seek social connectedness on social media to boost and alleviate their loneliness. Social media platforms allow people to express their thoughts, experiences, beliefs, and emotions. Prior studies on mental health  from social media have focused on symptoms, causes, and disorders. Whereas an initial screening of social media content for interpersonal risk factors and low self-esteem may raise early alerts and assign therapists to at-risk users of mental disturbance. Standardized scales measure self-esteem and interpersonal needs from questions created using psychological theories. In the current research, we introduce a psychology-grounded and expertly annotated dataset, \textbf{LoST}: \textbf{Lo}w \textbf{S}elf es\textbf{T}eem, to study and detect \textit{low self-esteem} on Reddit. Through an annotation approach involving checks on coherence, correctness, consistency, and reliability, we ensure gold-standard for supervised learning. We present results from different deep language models tested using two data augmentation techniques. Our findings suggest developing a class of language models that infuses psychological and clinical knowledge.   
\end{abstract}


\section{Introduction}

In the first year of the COVID-19 pandemic, the prevalence of anxiety and depression has increased by 25\%. Yet, according to World Health Organization (WHO) reports\footnote{\url{https://www.who.int/news/item/02-03-2022-covid-19-pandemic-triggers-25-increase-in-prevalence-of-anxiety-and-depression-worldwide}}, many of these cases have gone undiagnosed. Interpersonal risk factors, including loneliness and low self-esteem, have impacted individuals' mental health, triggering sub-clinical depression that worsens into clinical depression if left untreated. A recent study uses the interpersonal requirements of belongingness and positive self-esteem to highlight the latent phase of  \textit{depression} to \textit{suicidal ideation}~\cite{levi2021interpersonal}. 
\begin{figure}
    \centering
    \includegraphics[width=0.40\textwidth]{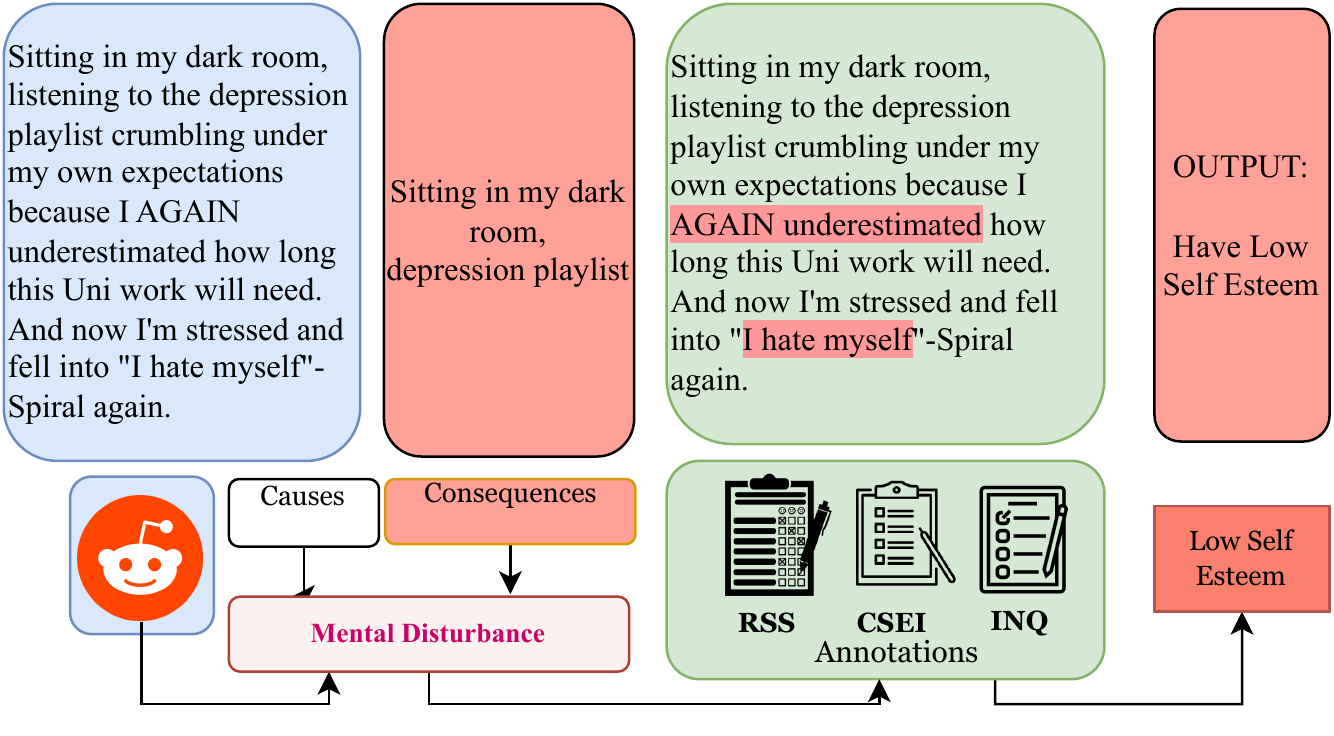}
    \caption{Overview of annotation scheme for the LoST dataset. Assign a specialist through early prediction of hallucinations of low self-esteem. Our task of classifying Reddit posts with low self-esteem facilitates the support for early control mechanisms on an increasing levels of depression and the severity of suicide.}
    \label{fig:1}
\end{figure}
Poor self-esteem is a common issue. Estimates of incidence in the general population go up to 85\%\footnote{\url{https://www.guttmanpsychology.com/2019/06/27/the-relationship-with-yourself/}}. Depression and the interplay between depression and low self-esteem can result in excessive stress, subpar performance under challenging circumstances, social anxiety, and reduced quality of life \cite{acarturk2009incidence}. Further, researchers illustrate low self-esteem as a major triggering point for increased risk of depression, anxiety, and suicidal ideation, affecting cognitive functioning, sleep quality, and overall well-being~\cite{korkmaz2019suicide}. Youth with multiple suicide attempts were likelier to have persistent suicide ideation, interpersonal struggles, feeling disconnected from others, and low self-competence~\cite{choi2013risk}. Past works demonstrate low self-confidence in people with low self-esteem who have a close association with social disengagement~\cite{watson2012rejection}.
%The \textit{low self-esteem} is a risk factor for developing social anxiety~. 
On the other hand, having strong or ``intact" self-esteem can be a barrier against the onset of mental illness. So, raising one's self-esteem seems like a sensible strategy for preventing and treating mental illness in the general population.  

A major challenge in the US is the overburdened healthcare system, which has forced users (early adults and teenagers) to see alternate avenues to meet their treatment needs.  For instance, Reddit has been the go-to anonymous social media platform for people to express their thoughts and beliefs easily without being judgmental of each other's experiences. Prior works demonstrate the potential to learn about social media users' mental health conditions using user-generated text and behavioral analyses of their social media activity~\cite{burke2010social}. In this work, we focus on detecting user-generated Reddit posts with low self-esteem to indicate the prospective presence of depression (see illustration in Figure~\ref{fig:1}). 

Advancements in natural language processing (NLP) for understanding the language of mental health on social media is significant traction, with a purpose to assist therapists and ensuring users at risk are recognized promptly~\cite{de2013predicting}~\cite{garg2023mental}. Datasets to support mental health research using NLP and artificial intelligence (AI) have focused more on statistical features and less on standardized scales or questionnaires (SCQ). Recently, Roy et al. and Zirikly et al. have demonstrated explainable detection of suicide and depression, respectively, by using datasets accompanied with SCQ \cite{zirikly2022explaining, roy2022process} Through the following post \textbf{P}, we present the benefit of having dataset annotated using SCQ:

\begin{quote}
    \textbf{P}: I \textcolor{red}{feel like a loser} because \textcolor{blue}{I do not have a group of friends}. I have friends but I \textcolor{blue}{do not have a group of them} to be with while I'm at college. I \textcolor{red}{feel like a loser} because I'm always alone. Unfortunately \textcolor{red}{I'm to blame} because I am extremely introverted and self conscious. \textcolor{blue}{I'm self conscious} to the point where I won't be friends with people because acquaintances mocked them frequently. \textcolor{blue}{I'm very lonely}. I began to \textcolor{blue}{use weed as a coping mechanism for my depression} and we all know how that has turned out. I'm very low risk because I believe that I have more to live for. I'm just \textcolor{blue}{tired of feeling alone}.
\end{quote}

The author's perception of being unsuccessful and mentally weak in post \textbf{P} suggests the presence of \textit{low self-esteem} in a given text. It exemplifies a recent model that emphasizes how self-esteem relates to perceptions of one's value concerning personal adequacy \cite{rimes2023low} and loneliness~\cite{uram2022still}. The highlighted phrases in \textbf{P} are the focused text spans which presents the warning signs of low self-esteem. These text spans presents psychological theories used by annotators in creating the proposed dataset. 

\paragraph{\textbf{Psychological Ground}} According to an American psychologist Cox's theory of the \textit{``hierarchy of human needs,"} having a high sense of self-worth is a fundamental need ~\cite{cox1987rich}. He distinguishes between two types of ``esteem": the \textit{need for respect from others} in the form of success, admiration, and recognition, and (ii) the need for respect from oneself in the form of self-love, self-confidence, skill, or aptitude. In this context, two professionals—a clinical psychologist and a social NLP researcher—worked together to design the annotations guidelines. To diagnose poor self-esteem, our professionals use SCQs, including Rosenberg's Self-esteem Scale (RSS), Coopersmith Self-Esteem Inventory (CSEI), and Interpersonal Needs Questionnaire (INQ-18).

\paragraph{\textbf{Our Contributions}} Our goal is to facilitate public health surveillance and health applications via releasing mental health data annotated from social media posts and analyzing current AI models for low self-esteem detection. To the best of our knowledge, the quantitative literature on \textit{low self-esteem and mental health} has no publicly available language resources due to the sensitive nature of the data. To this end, our \textbf{\textit{contributions}} can be summarized as follows:

\begin{itemize}
   
    \item We construct and publicly release LoST: \textbf{Lo}w \textbf{S}elf-es\textbf{T}eem, a new psychology-grounded dataset of 3,251 Reddit posts to facilitate social computing in mental health.
    \item We create a robust dataset considering FAIR principles to facilitate reliability and reusability ~\cite{dunning1970fair}.
    \item We experimented with deep language models as classifiers for low self-esteem detection and established them as baselines to identify challenges toward better AI solutions.

\end{itemize}

% \section{Related Work}

\section{Dataset}
We adhere to the ethical considerations for constructing and releasing the LoST dataset in the public domain. In this section, we first discuss the construction of a corpus (see Section~\ref{corpusconstruction}) followed by the expert-driven annotation scheme (see Section~\ref{annotationscheme}).
Our experts, a senior clinical psychologist and a social NLP researcher, frame \textit{annotation scheme} through extensive discussions on guidelines and perplexities, train three postgraduate students for eight hours and employ them on the annotation task. In this section, we further examine the annotation task's coherence, correctness, consistency, and reliability (see Section~\ref{annotationtask}).

\subsection{Corpus Construction}
\label{corpusconstruction}
We extract $200$ Reddit posts per day with subreddits \texttt{r/depression} and \texttt{r/SuicideWatch} from 2 December 2021 to 4 January 2022 through the Python Reddit API Wrapper (PRAW) API\footnote{\url{https://praw.readthedocs.io/en/stable/}}. We filter the \textit{candidate posts} with three major criteria: 
\begin{enumerate}
    \item We keep the body with length $>$ 0, and do not release information about any metadata to adhere to ethical constraints.
    \item We first identify the supportive statements, and remove the posts that do not reflect the mental disturbance. For example, we remove the following posts:
    \begin{quote}
       \textbf{P1}: Mental health is a very important part of one's life. One should focus on their mental and emotional wellbeing.\\
        \textbf{P2}: We encourage and welcome all our friends with mental disorders to join hands with us on our platform and lets make our journey beautiful and courageous.
    \end{quote}
    The generic nature of the Reddit posts \textbf{P1} and \textbf{P2} does not convey any significant information about individuals who are at-risk. 
    
    \item We remove the posts which reflect the intent of self-harm and suicidal tendencies without any contextual information about \textit{cause}~\cite{garg2022cams} and \textit{consequences}~\cite{dunn1959high}. For example, consider the post below:
    \begin{quote}
        \textbf{P3}: I am done with my life. 
        I don't want to live anymore.
    \end{quote}
    The given post \textbf{P3} highlights the user's suicidal ideation but there is no context (representing cause or consequence of suicidal tendency).
    
\end{enumerate}

We obtained a total of $4,357$ \textit{candidate posts}. The length of about 25\% of the posts exceed $300$ words. As our psychology-driven task of identifying low self-esteem is highly complex, we simplify it by filtering the \textit{candidate posts} with a total number of words more than $300$. We obtain a final corpus of $3,251$ posts, deploying it for the annotations task. 

\subsection{Annotation Scheme}
\label{annotationscheme}

A highly subjective and complex problem of detecting the low self-esteem in a given text may induce errors with \textit{naive} judgment. To mitigate this problem, we built a team of a clinical psychologist (reading between the lines to understand the psychological perception of the human mind) and a social NLP expert (text-based marking for outstanding AI models). To mitigate the trade-off between clinical psychology and NLP domain, experts suggest fine-grained guidelines to mark low self-esteem as a latent feeling of self-doubt, worthlessness, and lack of confidence. To facilitate the annotation task, our experts frame the annotation scheme, leveraging on two research questions: (i) ``RQ1: \textit{Does the text contain indicators of low self-esteem which alarms suicidal risk or self-harm in a person?}", and (ii) ``RQ2: \textit{What should be the extent to which annotators are supposed to read in-between-the-lines for marking the presence or absence of low self-esteem}". In this section, we discuss the annotation guidelines (see Section~\ref{guidelines}) and perplexity guidelines (see Section~\ref{perplexity}) to ensure future \textit{coherence} during the annotation task.
  
\subsubsection{Annotation Guidelines}
\label{guidelines}
Prior work examines self-esteem in terms of the \textit{self-worth} suggesting three dimensions of self-esteem: \textit{worth-based, efficacy-based, and authenticity-based esteem}~\cite{stets2014self}. With this background knowledge and comprehensive literature survey, 
our experts follow SCQs such as RSS \cite{rosenberg1965rosenberg}, a well-established questionnaire for detecting low self-esteem, to frame the annotation guidelines. RSS is a \texttt{ten-item scale} with items answered on a four-point scale—from \textit{strongly agree} to \textit{strongly disagree}. Among these ten items, five items are positively worded statements, and the remaining five are negatively worded statements to measure global self-worth with a score of 0-40. We also consider the adult version of the CSEI, an 58-item inventory designed for an assessment of an individual's global self-esteem~\cite{adair1984coopersmith}. Furthermore, our experts consider an 18-item INQ to make the annotation guidelines even more mature. The experts annotate 50 data-points of the corpus, in isolation, using a fine-grained guidelines to avoid biases. We discover 40\% of possible dilemmas in annotation task due to the subjective nature of the task. To resolve this problem, we identify the problems with annotation guidelines, and address these gaps with perplexity guidelines.  

\subsubsection{Perplexity Guidelines}
\label{perplexity}
We propose perplexity guidelines to simplify the task and facilitate future annotations. We observe two major confusions:
% \muskan{Rephrase language below}
\begin{enumerate}
    \item \textbf{Low Self-esteem in the Past}: To check if the condition of a person with low self-esteem is still an alarming prospect of self-harm or suicidal risk. Consider a post \textbf{A1} given below:
    
    \begin{quote}
   \textbf{A1}: I used to be boring and unattractive when no one used to like me, especially before Christmas but today I am celebrating this 
   New Year with my friends where everyone likes me.
    \end{quote}

    We frame rules to handle such contradictory statements (as shown in \textbf{A1}) that contain text-spans indicating both: (i) \textit{boring and unattractive}: an unpleasant perception about oneself before Christmas, and (ii) \textit{liking me}: people liking the author. A clear understanding of knowing oneself through public opinion is exemplified in the post, suggesting the temporary acceptance. The arguments of our experts argues the presence of low self-esteem in the user's perception which may reiterate after the temporary celebration. Thus, our experts recommend such contradictions as reflecting low self-esteem.
    
    \item\textbf{Ambiguity with \textit{Social Experiences}}: Sometimes, the unjust behaviour or biases of society makes a person mentally disturbed. In our work, we keep our guidelines supporting the detection of the low esteem within oneself as compared to the public opinion. Consider the post \textbf{A2} below:

    \begin{quote}
   \textbf{A2}: They don't like me at all. My friends are
   mean to me. They told me that they are not
   having any celebrations, but in reality,
   they were celebrating in amusement park.
   I don't care about what they think, but now its all boring.
    \end{quote}

    In \textbf{A2}, the author is feeling alone because of some subjective biases. Although public opinion may seed the feeling of interpersonal risk factors within an individual, such texts are not marked as the presence of LoST if they are not explicitly stated in a given text, as we choose not to make any assumptions.\footnote{We acknowledge that we make contextual observations by reading between the lines, but no emotion-driven assumptions to enable unbiased classification.} As a result, we observe the user expressing the feeling of \textit{lonesomeness} and \textit{alienation} which is clearly not rooted in low \textit{self}-esteem. Thus, we mark a given text having no signs of low self-esteem. 
\end{enumerate}

\subsection{Annotation Task}
\label{annotationtask}

% \muskan{Rephrase below}
We employ three postgraduate students for manual annotations, and experts train them for eight hours through \textit{annotation scheme}, ensuring their \textbf{coherence}. After three successive trial sessions to annotate 40 samples in each round, we ensure their \textbf{correctness} for alignment and understanding of the task to facilitate coherent annotations. %We found a slight lack of synchronization among annotators and acknowledge that the low levels of inter-annotator agreement are a well-known problem in emotion-based subjective studies, where lower agreement scores are reported ~\cite{tsakalidis2018building}.

All three students were made to sit in three different rooms and annotate the files individually to avoid any biases. Each student was given a task to annotate 50 samples per day to maintain their \textbf{consistency} for 66 weekdays. We further validate all three annotated files using the \textit{Fleiss' Kappa inter-observer agreement} study, where $\kappa$ is calculated as $78.52\%$, ensuring the \textbf{reliability} of our judgment followed by the experts' validation. We obtain final annotations based on the \textit{majority voting} mechanism. %We describe the statistics of the LoST (see Table~\ref{tab:2}). %In this section, we further examine the frequency of \textit{n-grams} and \textit{keyword extraction mechanisms} to understand the versatility of the words and key terms distributed across both classes. 
We Justify the findability and reusability of LoST through FAIR principles.% We explain the \textit{\textit{original data}} of the imbalanced dataset and perform the \textit{data augmentation} to enhance a number of the data points for the low represented class of the LoST. 

\subsection{FAIR Principles}
The FAIR guiding principle increases the \textbf{F}indability, \textbf{A}ccessibility, \textbf{I}nteroperability, and \textbf{R}eusability of the dataset to emphasize the machine actionability due to increasing reliance on computational systems to facilitate future studies~\cite{wilkinson2016fair}. 

\begin{enumerate}
    \item \textbf{Findable}: The LoST dataset contains the \textit{text} and the \textit{label} for each of the $3,251$ data-points. We release this dataset as the first version of our dataset as $LoST.v1$ at Github.\footnote{\url{https://github.com/drmuskangarg/LoST}}   % s.v1openly searchable with a persistent link that is uniquely attached to each specific dataset.

    \item \textbf{Accessible}: The LoST dataset is available in the comma-separated format on the Github. We plan to expand and update our dataset with explainable and contextual information in upcoming versions.

    \item \textbf{Inter-operable}: The LoST data is consistently structured and described, both syntactically and semantically, to support other major tasks of mental health analysis such as mental health triaging. 

    \item \textbf{Reusable}: We sufficiently annotate the dataset with $3,251$ instances for a binary classification problem to facilitate its re-usability. % so machine and human users can determine fit-for-purpose in the context of their analysis. 
    \end{enumerate}
    
\noindent To adhere to the ethical constraints of \textit{privacy, safety, and accountability}, we do not make any metadata available in the public domain, yet our dataset can be used effectively and opens up new research directions in NLP-centered mental health analysis.

\begin{table}[!ht]
\centering
\scriptsize
\caption{Comparison of the baseline methods with the Precision (P), Recall (R), F1-score (F1), and Accuracy, are averaged over 10-fold cross validation. Absent: Absence of Low Self-Esteem, Present: Presence of Low Self-Esteem. The second column specifies Composition (C) as Original data (O) or Augmented data (A)}
\label{resultss}
\begin{tabular}{cc|cccccc|c}
 \toprule[1.5pt]
\textbf{Model}   & \textbf{C}   & \multicolumn{3}{c}{\textbf{Absent}}  & \multicolumn{3}{c}{\textbf{Present}} & \textbf{Accuracy} \\    & & \textbf{P} & \textbf{R} & \textbf{F} & \textbf{P} & \textbf{R} & \textbf{F} & \\ \midrule 
LSTM  & O & 0.81 & 0.92 & 0.86 & 0.50 & 0.26 & 0.34  & 0.77  \\
& A & 0.84 & 0.72 & 0.78 & 0.72 & 0.84 & 0.78&  0.78   \\
\midrule
GRU  & O &0.81 & 0.91 & 0.86  & 0.48 & 0.27 & 0.35 & 0.76 \\ 
& A & 0.84 & 0.82 & 0.83 &  0.80 & 0.82 & 0.81  & 0.82\\
\midrule
BERT  & O & 0.89 & 0.86 & 0.88 &  0.57 & 0.62  &    0.60 &  0.81 \\
& A &  0.96 & 0.84 & \textbf{0.89}  & 0.78 & 0.94 & \textbf{0.85}   &  \textbf{0.88}\\
\midrule
RoBERTa & O &  0.92 & 0.86 & \textbf{0.89} & 0.51 & 0.67 & 0.58 &  \textbf{0.82}\\
& A & 0.90 & 0.83 & 0.86 & 0.80 & 0.88 & 0.84   &0.85 \\
% \midrule
% DistilBERT  & O & 1.00 & 0.76 & 0.86  & 0.00 & 0.00 & 0.00& 0.76\\
% & A & 0.86 & 0.84 & 0.85 &  0.82 & 0.84 & 0.83  & 0.84\\
\midrule
XLNet & O &  0.86 & 0.89 & 0.87 & 0.65 & 0.59 & \textbf{0.62} &  0.81  \\
& A & 0.95 & 0.80 & 0.87 &  0.74 & 0.93 & 0.82 & 0.85\\

% GloVe +LSTM & 0.64 &	0.38 & 0.47 & 0.67 & 0.86 &	0.76 & 0.67\\
% GloVe+BiLSTM & 0.65 &	0.48 & 0.56 &	0.71 &	0.83 &	0.76 &	0.69\\

% GloVe + GRU & 0.62 &	0.60 &	0.61 &  0.74 & 0.76 &	0.75&	0.69\\
% GloVe+ BiGRU & 0.64 & 0.54 & 0.58 & 0.72 &	0.80 &	0.76 & 0.70 \\
% % ELMo  & 0.59     & 0.57      & 0.60    \\
% BERT & 0.52 & 0.79 & 0.63 & 0.77 & 0.48 & 0.59  & 0.61      \\
% RoBERTa & 0.89 & 0.14 & 0.24 & 0.62 & 0.99 & 0.76 & 0.64      \\
% ALBERT & 0.63     & 0.58      & 0.61           \\\midrule \midrule
% DistilBERT &           &        &         \\
% XLNET      &           &        &         \\  
\bottomrule[1.5pt]
\end{tabular}

\end{table}

\section{Experiments and Evaluation}
In this section we perform experiments with the LoST dataset over six different classifiers and evaluate their performance of low self-esteem detection with precision, recall, F1-score and accuracy. To measure the impact of imbalanced dataset, we use Matthews Correlation Coefficient (MCC) which range from $-1$ to $+1$ where the values closer to $0$ suggests randomization, the values closer to $+1$ suggests the extent of perfection in prediction and the values closer to $-1$ shows the poor models~\cite{chicco2020advantages}. The MCC on \textit{original data} of the dataset suggests the need of adding more samples for less represented class. We employ two data augmentation techniques and experiment all baselines on the resulting dataset. 

\paragraph{\textbf{Classifiers}}

We perform extensive analysis to build baselines and highlight their limitations. We considered sequence to sequence and attention-based models as considerable baselines for the task. The following classifiers
have shown state-of-the-art performances in their respective studies: 
(i) \textit{Recurrent Neural Networks} (RNN) (LSTM~\cite{bai2018text}, GRU~\cite{cho2014properties}), (ii) \textit{Pre-trained transformer based models} such as BERT~\cite{martinez2021bert}, RoBERTa~\cite{murarka2020detection}, and XLNet~\cite{alshahrani2020identifying}.%  to obtain features and feed them to linear classifier for performance evaluation.
    % \end{itemize}

\begin{enumerate}
    \item \textbf{LSTM:} %The Long Short-Term Memory networks (LSTM) takes a sequence of data as input and make predictions at individual time steps of the sequential data. 
    We apply the LSTM model for classifying the texts with low self-esteem from those data samples which do not.
    \item \textbf{GRU:} %A Gated Recurrent Unit (GRU) intends to use connections between the sequence of nodes to resolve the vanishing gradient problem. 
    Since the textual sequences in the LoST present a mixed context, GRU's non-sequential nature provide improvement over LSTM. 

     %We apply GRU model which undergo faster training of training data as compared to that of LSTM due to less number of parameters.
     
    %to perform classification task and vanishes the gradient problem by using less training parameter and therefore uses less memory utilization. Thus, GRU executes faster than LSTM.
    \item \textbf{BERT:} %BERT’s key technical innovation is applying the bidirectional training of transformer, a popular attention model for language modeling. 
    We use the \texttt{bert-base-uncased} model for our task. The attention feature in BERT captures better context on the LoST dataset than GRU\footnote{\url{https://huggingface.co/bert-base-uncased}; GRU is undirectional sequence to sequence model}. 
    \item \textbf{RoBERTa:}  RoBERTa model is built on BERT by modifying the key hyperparameters, removing the next-sentence pretraining objective and training with larger mini-batches and learning rates. We use \texttt{roberta-base} model for our task.\footnote{\url{https://huggingface.co/roberta-base}}
   % \item \textbf{DistilBERT:} DistilBERT is a small, fast, cheap and light Transformer model based on the BERT architecture. Knowledge distillation is performed during the pre-training phase to reduce the size of a BERT model by 40\%. We used \\\texttt{distilbert-base-uncased} model for our task.\footnote{\url{https://huggingface.co/distilbert-base-uncased}}
    \item \textbf{XLNet:} XLnet is an extension of the Transformer-XL model pre-trained using an autoregressive method to learn bidirectional contexts by maximizing the expected likelihood over all permutations of the input sequence factorization order. We used \texttt{xlnet-base-cased} model for our task.\footnote{\url{https://huggingface.co/xlnet-base-cased}}
\end{enumerate}

\paragraph{\textbf{Experimental Setup}.}
For consistency, we used the same experimental settings for all models with 10 fold cross-validation. All our results are reported as the average across all folds. A varying lengths of posts are padded and trained for 150 epochs with early stopping with a patience of 10 epochs. Thus, we set hyperparameter for our experiments with transformer-based models as $H$ = $200$, $O$ = Adam, learning rate = 1$\times 10^{-5}$, and batch size $16$.%, and $10$ epochs. %We further regularize LSTM and GRU with kernel regularization and bias regularization of 1e-4 learning rate.
% \subsection{Performance Measures}
%\subsection{Evaluation Metrics}

% \subsection{Classifiers}
\subsection{Experimental Results}
\subsubsection{Original data}
Table I contains the classification performance of low self-esteem detection. The results suggest the average performance with the low values of MCC, and the need of reducing data imbalancing. We postulate this average performance due to the incapability of capturing contextual information by existing classifiers. Among the existing ones, RoBERTa outperforms all the other models for \textit{original data} of the model and counterparts with 82\% of the accuracy. We further determine the values of MCC for classifiers on the \textit{original data} of LoST that is observed as an average of +0.20 for RNN models and +0.48 for pre-trained transformer based models, suggesting the randomized prediction, especially for RNN models. We employ data augmentation techniques to reduce the data imbalancing.

%of LoST that is observed as an averaged of $+0.20$ for RNN models and $+0.48$ for pre-trained tranformer based models, suggesting the randomized prediction, especially for RNN models. We employ the data augmentation techniques to reduce the data imbalancing. %Furthermore, low MCC scores suggests the need of comparatively balanced dataset.

\subsubsection{Augmentated data}
As observed before, the ratio of data-points for $presence : absence$ is $22.4 : 77.6$ which approximates to 1 : 3. To handle this problem of imbalanced dataset, we multiply the positive data-points with two data augmentation approaches: % The low number of samples with presence of low self-esteem supports the existing psychology-grounded theory of XXX. The XXX theory states that XXXX suggesting XXXX. 
% To handle this problem of imbalanced dataset, we multiply the positive data-points with two data augmentation approaches:
\begin{enumerate}
    \item \textbf{Easy Data Augmentation (EDA):} We employ an EDA method for generating an additional corpus of $729$ samples~\cite{wei2019eda}. Originally EDA comprises Synonym Replacement (SR), Random Insertion (RI), Random Swap (RS), Random Deletion (RD). We found a dysfunctional mechanism of RS as the lexical sequence deformation. Thus, we use SR, RD with $20\%$, followed by RI. Consider an example $E$ below:
    \begin{quote}
        \texttt{E}: I am good for nothing and thus, jobless since two years.\\
        \texttt{Augmentation}: I am good for nothing \textbf{and}$\leftarrow$ \textcolor{red}{RD} thus, \textbf{jobless}$\leftarrow$ \textcolor{red}{umemployed SR} since \textbf{two}$\leftarrow$ \textcolor{red}{RD} \textcolor{red}{few $\leftarrow$ SR} years \textcolor{red}{now $\leftarrow$ RI}.\\
        \texttt{Augmented text (E}): I am good for nothing thus, unemployed since few years now.
    \end{quote}
    \item \textbf{Back Translation (BT):} We use the French language for BT~\cite{sennrich2015improving} to add $729$ samples for positive data points of LoST dataset. Consider the example $E$ given above and following conversions with back translation:
    \begin{quote}
        \texttt{French Translation}: Je suis bon à rien et donc sans travail depuis deux ans.\\
        \texttt{Augmented text (E}): I'm good for nothing and therefore out of work for two years.
    \end{quote}
\end{enumerate} 
We obtain $3$ times $729$ samples as a resulting set of data-points for label 1. Our initiative with data-augmentation reduces the imbalancing of LoST. We further perform our experiments with both \textit{\textit{original data}} and \textit{augmented data}. We observe the improvement in MCC for RNN models (with an average of $+0.64$) and pre-trained transformer-based models (with average of $+0.73$), suggesting reliable inferences. For all the models, we observe improved accuracy and F1-score with augmented data in comparison of \textit{original data} of LoST. The pre-trained model BERT outperforms all the baselines, suggesting the need of more explainable and accountable models. 

\subsection{Discussion}
Table \ref{resultss} reveals that, in contrast to posts with self-esteem signals, our baselines are confident in categorizing posts with no signals of self-esteem. This suggest room for improvement for AI and NLP to develop novel algorithms for detecting warning signs of low self esteem. 
We considered as baselines, a diverse class of language models (base, not fine-tuned): Sequence to sequence, attention-based, knowledge distillation, and autoregressive which have been predominantly used in prior literature~\cite{gaur2022knowledge,futami2020distilling}. All the models built over augmented data saw an averaged 7\% gain in accuracy, and such a consistency informs dependability of the dataset. Irrespective of the size of the model, the BERT and RoBERTa language models consistently outperformed, highlighting the significance of attention. We observe that by adding external knowledge, such the knowledge in SCQs, autoregressive models (like XLNET) can outperform BERT/RoBERTa. %The strange behaviour of DistilBERT for Original data and excellent performance with augmented data shows the integrity of models built over augmented data. 

% Quantitative Improvement
%The best performance is observed with accuracy of BERT and RoBERTa for augmented data and original data, respectively.
%The augmented dataset supports better model development with higher MCC values, indicating the reliability of dataset. The XLNet gives 6.8\% better F1-score than RoBERTa for original data. DistilBERT shows poor performance due to highly imbalanced nature of the data. We observe the consistency in improvements for models built over augmented data as compared to the original data.  
% Composition, kaunsi consistently outperform and why
% Accuracy -- imbalance nature
% Finsdings -- 
%Irrespective of size of the model, BERT and RoBERTa language model outperformed consistently, signifying the importance of attention. %is important, 
%The statistical permutation or combination-type models like XLNet or knowledge distilation models are incapable and such intrinsic methods in language models should be supported by external knowledge, such as the information in SCQs. 

%The better performance of classifiers with augmented datasets are due to the repetitive nature of data points. 
We recommend infusing clinically grounded knowledge to build more informative models and test the classifiers with human-generated explanations, keeping them as an open research direction for future developments. We recommend the use of SCQs such as RSS, CSEI, and INQ-18 to build a lexicon on the experts' recommendations and infuse it as an external knowledge to build more contextualized models. We further suggest developing consequence-preserving data augmentation techniques on the \textit{original data} of LoST for better representation. The practical considerations for this task is applied to problems with work-life and abusive relationships. For instance,
the research community has witnessed a surge in the recent issue of job layoffs since the COVID-19 pandemic~\cite{el2023first}, highlighting the problems of low self-esteem in affected people. Other than jobs and careers, poor performance in school, hallucinations of unattractive personality among friends and family, and failed relationships are other problems caused by low self-esteem. An early check on such situational mental disturbance may prevent chronic disease of clinical depression and suicidal ideation.

% \section{New Frontiers}\textbf{}

\section{Conclusion}
We present LoST, a SCQ-informed gold standard dataset of $3,251$ Reddit Posts for low self-esteem to facilitate social computing in the mental health domain. Our experts-driven annotation scheme and the annotation task enable a coherent, complete, consistent, and reliable dataset. We developed binary classifiers and observed the best performance of RoBERTa (with 82\% accuracy on \textit{original data} of data) and BERT (with 88\% accuracy on augmented data). The implications of this work include the potential to improve public health surveillance and health applications that rely on automatically identifying con- sequences in the posts in which users describe their mental health issues. In future work, we plan to enhance the LoST dataset with more samples and additional labels of contextual explanations (similar to CAMS~\cite{garg2022cams}). We further plan to develop new models tailored explicitly to low self-esteem by infusing knowledge through clinical psychology-grounded lexicons.

%We present LoST, a new dataset of $3,251$ Reddit Posts to detect low self-esteem, facilitating its reusability. Our experts-driven annotation scheme and the annotation task suggest a coherent, complete, consistent, and reliable dataset. We further employ baseline methods as binary classifiers and observe the best performance of RoBERTa (with 82\% accuracy) and BERT (with 88\% accuracy) for \textit{\textit{original data}} and \textit{augmented dataset}, respectively. The implications of this work include the potential to improve public health surveillance and other health applications that rely on automatically identifying consequences in the posts in which users describe their mental health issues. In future work, we plan to enhance the \textit{LoST dataset} with more samples and the current version of $LoST.v1$ with additional labels of contextual explanations. We further plan to develop new models tailored explicitly to low self-esteem by infusing knowledge through clinical psychology-grounded lexicons.% detection. \muskan{Add the need of understanding the perspective of a user for KiL through the higher layers of NLP analysis discourse analysis and pragmatics. }

% The explainable AI models developed for lonesomeness classification is applicable to determine social media at-risk users with suicidal prospects. The consequence of loneliness derives the need of early mitigation of deteriorating mental health. 

\section*{Ethics and Broader Impact}
% \muskan{\url{https://arxiv.org/pdf/2106.01702.pdf}}

% \muskan{Rephrase below}

% \vspace{-0.2cm}\section{Ethics and Broader Impact}
Social media data is often sensitive, especially when the data is related to mental health~\cite{benton2017ethical}. Our LoST dataset contains only publicly available posts, and no user's metadata is made available as we are committed to the
ethical practices of protecting the privacy
and anonymity of the users~\cite{henderson2018ethical}. The examples shown in this paper are anonymized, obfuscated, and paraphrased to prevent misuse. Thus, this study does not require any ethical approval. Due to the subjective nature of our task, we expect some biases in our annotations. We clearly define the annotation scheme and other instructions to address these concerns. Due to the high inter-annotator agreement (\(\kappa\) score), we are confident that most annotations are correctly assigned to the data. We release the LoST dataset to accelerate the research and development applicable to the screening tasks and mental health triaging. The dataset and the source code to reproduce the baseline results are available on GitHub.\footnote{\url{https://github.com/drmuskangarg/LoST}} Clearly, machine learning predictions cannot replace professional mental health diagnostics, counseling, or therapy. As shown in our evaluation, their accuracy and trustworthiness remain insufficient for such purposes~\cite{nicholas2020ethics}. We assume all posts to be a genuine expression of users' experiences and that they are not manipulative. 

\section*{Acknowledgement}
We extend our sincere acknowledgement to the postgraduate student annotators, Ritika Bhardwaj, Astha Jain, and Amrit Chadha, for their diligent efforts in the annotation process. We express our gratitude to Veena Krishnan, a senior clinical psychologist, and Ruchi Joshi, a rehabilitation counselor, for their unwavering support throughout the project. This project was partially supported by NIH R01 AG068007. We thank Surjodeep Sarkar for proofreading this work. %\textcolor{red}{ Please write acknowledgement}

\bibliographystyle{splncs04}
\bibliography{references}

\begin{thebibliography}{10}
\providecommand{\url}[1]{\texttt{#1}}
\providecommand{\urlprefix}{URL }
\providecommand{\doi}[1]{https://doi.org/#1}

\bibitem{acarturk2009incidence}
Acarturk, C., Smit, F., De~Graaf, R., Van~Straten, A., Ten~Have, M., Cuijpers,
  P.: Incidence of social phobia and identification of its risk indicators: a
  model for prevention. Acta Psychiatrica Scandinavica  \textbf{119}(1),
  62--70 (2009)

\bibitem{adair1984coopersmith}
Adair, F.L.: Coopersmith self-esteem inventories. Test critiques  \textbf{1},
  226--232 (1984)

\bibitem{roy2022process}
Roy~et. al., K.: Process knowledge-infused learning for suicidality assessment
  on social media. arXiv preprint arXiv:2204.12560  (2022)

\bibitem{gaur2022knowledge}
Gaur~et al., M.: Knowledge-infused learning: A sweet spot in neuro-symbolic ai.
  IEEE IC  (2022)

\bibitem{henderson2018ethical}
Henderson~et al., P.: Ethical challenges in data-driven dialogue systems (2018)

\bibitem{alshahrani2020identifying}
Alshahrani, A., Ghaffari, M., Amirizirtol, K., Liu, X.: Identifying optimism
  and pessimism in twitter messages using xlnet and deep consensus. In: 2020
  International Joint Conference on Neural Networks (IJCNN). pp.~1--8. IEEE
  (2020)

\bibitem{bai2018text}
Bai, X.: Text classification based on lstm and attention. In: 2018 Thirteenth
  International Conference on Digital Information Management (ICDIM). pp.
  29--32. IEEE (2018)

\bibitem{benton2017ethical}
Benton, A., Coppersmith, G., Dredze, M.: Ethical research protocols for social
  media health research. In: Proceedings of the first ACL workshop on ethics in
  natural language processing. pp. 94--102 (2017)

\bibitem{burke2010social}
Burke, M., Marlow, C., Lento, T.: Social network activity and social
  well-being. In: Proceedings of the SIGCHI conference on human factors in
  computing systems. pp. 1909--1912 (2010)

\bibitem{chicco2020advantages}
Chicco, D., Jurman, G.: The advantages of the matthews correlation coefficient
  (mcc) over f1 score and accuracy in binary classification evaluation. BMC
  genomics  \textbf{21},  1--13 (2020)

\bibitem{cho2014properties}
Cho, K., van Merri{\"e}nboer, B., Bahdanau, D., Bengio, Y.: On the properties
  of neural machine translation: Encoder--decoder approaches. In: Proceedings
  of SSST-8, Eighth Workshop on Syntax, Semantics and Structure in Statistical
  Translation. pp. 103--111 (2014)

\bibitem{choi2013risk}
Choi, K.H., Wang, S.M., Yeon, B., Suh, S.Y., Oh, Y., Lee, H.K., Kweon, Y.S.,
  Lee, C.T., Lee, K.U.: Risk and protective factors predicting multiple suicide
  attempts. Psychiatry research  \textbf{210}(3),  957--961 (2013)

\bibitem{cox1987rich}
Cox, R.: The rich harvest of abraham maslow. Motivation and personality pp.
  245--271 (1987)

\bibitem{de2013predicting}
De~Choudhury, M., Gamon, M., Counts, S., Horvitz, E.: Predicting depression via
  social media. In: Proceedings of the International AAAI Conference on Web and
  Social Media. vol.~7 (2013)

\bibitem{dunn1959high}
Dunn, H.L.: High-level wellness for man and society. American journal of public
  health and the nations health  \textbf{49}(6),  786--792 (1959)

\bibitem{dunning1970fair}
Dunning, A., De~Smaele, M., B{\"o}hmer, J.: Are the fair data principles fair?
  International Journal of digital curation  \textbf{12}(2),  177--195 (1970)

\bibitem{el2023first}
El-Deeb, A.: The first tech layoff wave after years of hypergrowth: How this
  affects the industry? ACM SIGSOFT Software Engineering Notes  \textbf{48}(1),
  ~4--5 (2023)

\bibitem{futami2020distilling}
Futami, H., Inaguma, H., Ueno, S., Mimura, M., Sakai, S., Kawahara, T.:
  Distilling the knowledge of bert for sequence-to-sequence asr. arXiv preprint
  arXiv:2008.03822  (2020)

\bibitem{garg2023mental}
Garg, M.: Mental health analysis in social media posts: A survey. Archives of
  Computational Methods in Engineering pp. 1--24 (2023)

\bibitem{garg2022cams}
Garg, M., Saxena, C., Krishnan, V., Joshi, R., Saha, S., Mago, V., Dorr, B.J.:
  Cams: An annotated corpus for causal analysis of mental health issues in
  social media posts. arXiv preprint arXiv:2207.04674  (2022)

\bibitem{korkmaz2019suicide}
Korkmaz, H., Korkmaz, S., {\c{C}}akar, M.: Suicide risk in chronic heart
  failure patients and its association with depression, hopelessness and self
  esteem. Journal of clinical neuroscience  \textbf{68},  51--54 (2019)

\bibitem{levi2021interpersonal}
Levi-Belz, Y., Aisenberg, D.: Interpersonal predictors of suicide ideation and
  complicated-grief trajectories among suicide bereaved individuals: A
  four-year longitudinal study. Journal of affective disorders  \textbf{282},
  1030--1035 (2021)

\bibitem{martinez2021bert}
Mart{\'\i}nez-Casta{\~n}o, R., Htait, A., Azzopardi, L., Moshfeghi, Y.:
  Bert-based transformers for early detection of mental health illnesses. In:
  Experimental IR Meets Multilinguality, Multimodality, and Interaction: 12th
  International Conference of the CLEF Association, CLEF 2021, Virtual Event,
  September 21--24, 2021, Proceedings 12. pp. 189--200. Springer (2021)

\bibitem{murarka2020detection}
Murarka, A., Radhakrishnan, B., Ravichandran, S.: Detection and classification
  of mental illnesses on social media using roberta. arXiv preprint
  arXiv:2011.11226  (2020)

\bibitem{nicholas2020ethics}
Nicholas, J., Onie, S., Larsen, M.E.: Ethics and privacy in social media
  research for mental health. Current Psychiatry Reports  \textbf{22}(12),
  ~1--7 (2020)

\bibitem{rimes2023low}
Rimes, K., Smith, P., Bridge, L.: Low self-esteem: A refined cognitive
  behavioural model. Behavioural and Cognitive Psychotherapy  (2023)

\bibitem{rosenberg1965rosenberg}
Rosenberg, M.: Rosenberg self-esteem scale. Journal of Religion and Health
  (1965)

\bibitem{sennrich2015improving}
Sennrich, R., Haddow, B., Birch, A.: Improving neural machine translation
  models with monolingual data. arXiv preprint arXiv:1511.06709  (2015)

\bibitem{stets2014self}
Stets, J.E., Burke, P.J.: Self-esteem and identities. Sociological perspectives
   \textbf{57}(4),  409--433 (2014)

\bibitem{uram2022still}
Uram, P., Skalski, S.: Still logged in? the link between facebook addiction,
  fomo, self-esteem, life satisfaction and loneliness in social media users.
  Psychological Reports  \textbf{125}(1),  218--231 (2022)

\bibitem{watson2012rejection}
Watson, J., Nesdale, D.: Rejection sensitivity, social withdrawal, and
  loneliness in young adults. Journal of Applied Social Psychology
  \textbf{42}(8),  1984--2005 (2012)

\bibitem{wei2019eda}
Wei, J., Zou, K.: Eda: Easy data augmentation techniques for boosting
  performance on text classification tasks. In: Proceedings of the 2019
  Conference on Empirical Methods in Natural Language Processing and the 9th
  International Joint Conference on Natural Language Processing (EMNLP-IJCNLP).
  pp. 6382--6388 (2019)

\bibitem{wilkinson2016fair}
Wilkinson, M.D., Dumontier, M., Aalbersberg, I.J., Appleton, G., Axton, M.,
  Baak, A., Blomberg, N., Boiten, J.W., da~Silva~Santos, L.B., Bourne, P.E.,
  et~al.: The fair guiding principles for scientific data management and
  stewardship. Scientific data  \textbf{3}(1), ~1--9 (2016)

\bibitem{zirikly2022explaining}
Zirikly, A., Dredze, M.: Explaining models of mental health via clinically
  grounded auxiliary tasks. In: CLPsych (2022)

\end{thebibliography}
\end{document}